\documentclass[11pt]{article}
\usepackage{arabtex}
\usepackage{utf8}
\setcode{utf8}

\usepackage[final]{acl} 

\usepackage[T1]{fontenc}
\usepackage[utf8]{inputenc}

\usepackage{times}
\usepackage{latexsym}
\usepackage{tabularx}
\usepackage{booktabs}
\usepackage{adjustbox}
\usepackage{multirow}
\usepackage{amsmath}
\usepackage{amssymb}
\usepackage{float}
\usepackage{longtable}
\usepackage{siunitx}
\usepackage{array}
\usepackage{csquotes}
\usepackage{afterpage}
\usepackage{microtype}
\usepackage{inconsolata}
\usepackage{array}
\usepackage{ragged2e}

\newcolumntype{L}{>{\raggedright\arraybackslash}X}
\newcolumntype{R}{>{\raggedleft\arraybackslash}X}

\usepackage{graphicx}
\usepackage{url}
\usepackage{placeins}
\usepackage{adjustbox}

\usepackage{comments}
\newauthor[Sh]{Simon}{yellow}  
\newauthor[Pb]{Pierrette}{red}     
\newauthor[Pa]{Perla}{green}  
\newauthor[Jm]{Johnny}{blue}  

\title{Aladdin-FTI @ AMIYA\\
\Large Three Wishes for Arabic NLP: Fidelity, Diglossia, and Multidialectal Generation}

\author{
  \textbf{Jonathan Mutal\textsuperscript{~\includegraphics[height=1.4em]{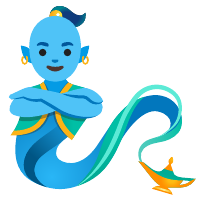}}}
  \textbf{Perla Al Almaoui\textsuperscript{~\includegraphics[height=1.4em]{man-genie.pdf}}}
  \textbf{Simon Hengchen\textsuperscript{~\includegraphics[height=1.4em]{man-genie.pdf},~\includegraphics[height=1.4em]{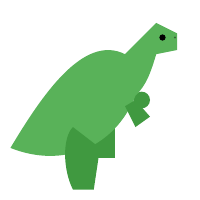}}}
  \textbf{Pierrette Bouillon\textsuperscript{~\includegraphics[height=1.4em]{man-genie.pdf}}}
\\
  \textsuperscript{~\includegraphics[height=1.2em]{man-genie.pdf}}Faculté de traduction et d'interprétation, Université de Genève\\
\textsuperscript{~\includegraphics[height=1.2em]{iguanodon_emoji.pdf}}{iguanodon.ai}\\[0.5em]
  \small{\textbf{Correspondence:} first.last@unige.ch}
}

\begin{document}

\maketitle
\begin{abstract}
%
%
Arabic dialects have long been under-represented in Natural Language Processing (NLP) research due to their non-standardization and high variability, which pose challenges for computational modeling. Recent advances in the field, such as Large Language Models (LLMs), offer promising avenues to address this gap by enabling Arabic to be modeled as a pluricentric language rather than a monolithic system.
This paper presents Aladdin-FTI, our submission to the AMIYA shared task. The proposed system is designed to both generate and translate dialectal Arabic (DA). Specifically, the model supports text generation in Moroccan, Egyptian, Palestinian, Syrian, and Saudi dialects, as well as bidirectional translation between these dialects, Modern Standard Arabic (MSA), and English. The code and trained model are publicly available.\footnote{Code: \url{https://github.com/drvenabili/mtfinetune_amiya}, models: \url{https://hf.co/collections/unige-fti/aladdin-fti-amiya}.}
\end{abstract}

\section{Introduction}
\label{sec:introduction}
The Thirteenth Workshop on NLP for Similar Languages, Varieties and Dialects (VarDial 2026)\footnote{\url{https://sites.google.com/view/vardial-2026}} introduces the \enquote{Arabic Modeling In Your Accent} (AMIYA) shared task \cite{robinson-etal-2026-amiya}, a benchmark designed to advance computational modeling of DA. 
The AMIYA shared task focuses on developing language models that capture the linguistic characteristics of spoken Arabic varieties. 
Such varieties remain under-represented in existing NLP research and resources \citep{harrat2019machine}, although there is a growing interest in studying dialectal varieties and more resources are being created (see e.g. \citet{al-haff-etal-2022-curras,atlasia2024terjamabench,almaoui-etal-2025-arabizi}).
In this evaluation campaign, systems are assessed on their ability to model DA with respect to dialectal fidelity, understanding, and generation quality using the AL-QASIDA benchmark \citep{robinson-etal-2025-al}.

\begin{figure}[H]
    \centering
    \includegraphics[width=1\linewidth]{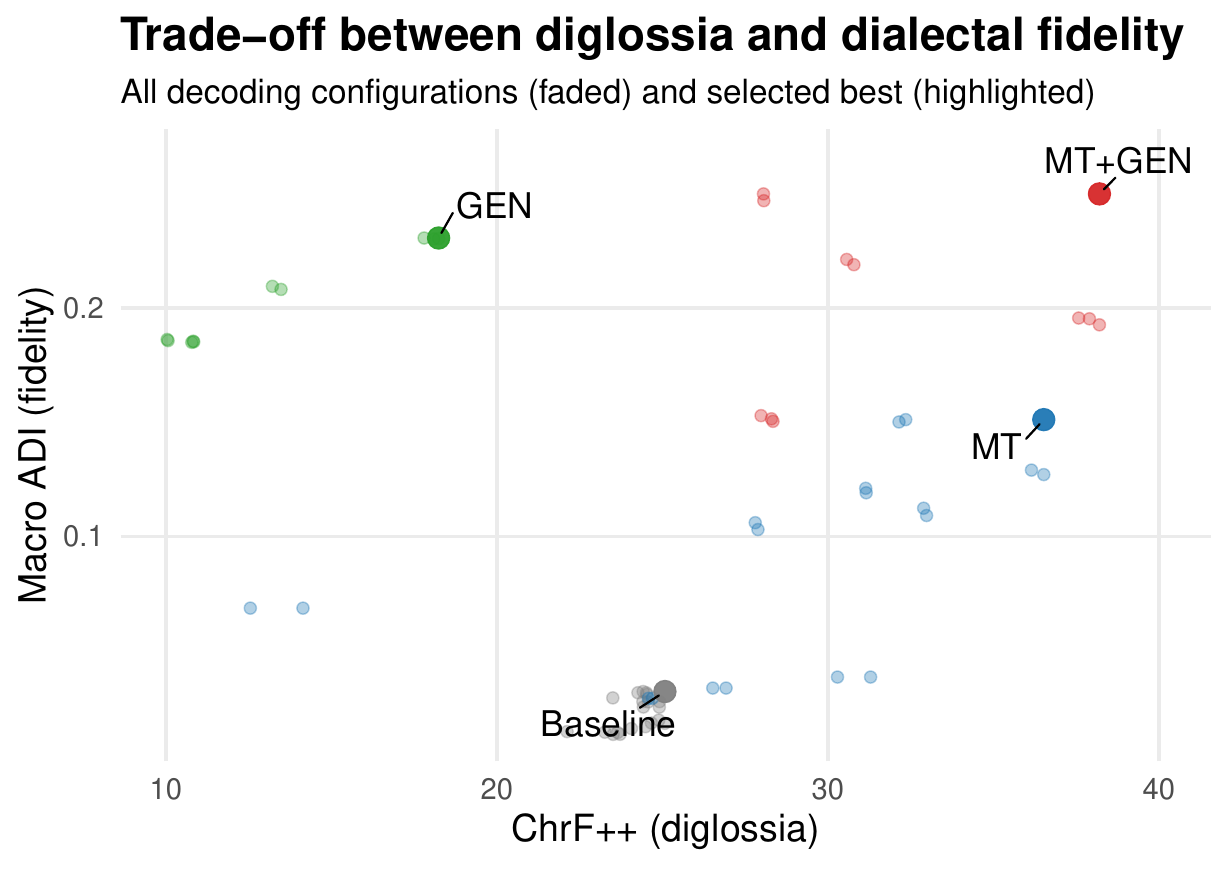}
    \caption{Trade-off between diglossia-sensitive translation accuracy (ChrF++) and dialectal fidelity (Macro ADI2). Each faded point corresponds to a decoding configuration (learning rate $\times$ checkpoint), while highlighted points indicate the best configuration selected per model. Instruction-based generation (GEN) favours dialectal fidelity at the expense of diglossia, whereas MT exhibits the opposite behaviour. The combined MT+GEN objective achieves the best overall, improving both fidelity and diglossia.}
    \label{fig:trade-off_mt_fidelity}
\end{figure}

This paper describes the participation of the team \textbf{\mbox{Aladdin-FTI ~\includegraphics[height=1em]{man-genie.pdf}}}, focusing on the closed data track, where the models are only fine-tuned on the official training data provided by the shared task organizers, without the use of additional external corpora. Our approach is based on translation and generation by combining two training objectives: (i) a translation objective aimed at reinforcing diglossic distinctions between MSA and DA, while also preserving semantic adequacy; and (ii) an instruction next-token generation objective designed to produce fluent and linguistically dialectal continuations from partial prompts. By jointly training with these objectives, we seek to have a balance between semantic adequacy and dialectal expressiveness~\citep{robinson-etal-2025-al}. We investigate the complementary roles of translation and generation in dialectal Arabic modeling along the following questions:

\begin{itemize}
    \item[\textbf{Q1}] What is the impact of  fine-tuning for translation on diglossia and dialectal fidelity across Arabic dialects?
    \item[\textbf{Q2}] What is the impact of instruction fine-tuning for next word generation on diglossia and dialectal fidelity across Arabic dialects?
    \item[\textbf{Q3}] What is the impact of both machine translation (MT) and instruction fine-tuning for next word generation on diglossia and dialectal fidelity across Arabic dialects?
\end{itemize}

We fine-tune a single large language model under different training settings and evaluate their impact on both diglossia and dialectal fidelity. Our results highlight their distinct yet complementary roles in DA generation.

Our contributions are the following:
\begin{itemize}
    \item We propose a joint training objective that combines machine translation and instruction-conditioned next-token generation for dialectal Arabic.
    \item This training enables smaller models to match or even outperform substantially larger baselines in modeling Arabic dialectal variation.
\end{itemize}

The remainder of this paper is organized as follows: first, Section~\ref{sec:related-work} reviews related work; next, Section~\ref{sec:methodology} presents the methodology; Section~\ref{sec:experimental_set-up} describes the experimental setup; Section~\ref{sec:results} reports the results; and finally, Section~\ref{sec:conclusion} concludes by discussing the limitations of the study.


%

\section{Related Work}
\label{sec:related-work}
%
Arabic has long been treated as a single homogeneous language, with the majority of resources, benchmarks, and models focusing almost exclusively on MSA. However, this perspective overlooks the deeply diglossic nature of Arabic-speaking communities, in which MSA is rarely a native language and is primarily used in formal and written contexts, while everyday communications take place in regionally and socially diverse dialects ~\citep{keleg-etal-2025-revisiting}. These dialects differ substantially from MSA in phonology, morphology, syntax, and lexicon, and each reflects the historical, cultural, and social identities of its speakers~\citep{yushmanov1961structure}.

This MSA-centric focus poses significant challenges for NLP systems, as models trained predominantly on MSA tend to normalize or suppress dialectal features when generating or translating dialectal text ~\citep{robinson-etal-2023-chatgpt}. This gap motivates research into methods that explicitly preserve dialectal characteristics in generated text while maintaining semantic adequacy.
\paragraph{Machine Translation for Dialectal Arabic}
Prior work has shown that MT provides a natural framework for modeling distinctions between MSA and DA, as translation objectives explicitly condition generation on a source sentence rather than on the target alone~\citep{habash-etal-2013-morphological,zbib-etal-2012-machine}. More recent efforts include the fine-tuning of the Kuwain-1.5B small language model for translation from 15 Arabic dialects into Modern Standard Arabic, which achieved high human-rated fluency scores in evaluation studies~\citep{hamed-etal-2025-lahjawi}, indicating improved generation quality for dialectal inputs.

Despite these advances, multiple studies report that dialectal MT systems often favour normalised outputs, exhibiting MSA lexical or morphosyntactic choices even when dialectal targets are explicitly specified~\citep{habash-etal-2013-morphological,bouamor-etal-2018-madar,robinson-etal-2025-al}. While translation-based approaches tend to preserve meaning effectively, they may under-represent dialect-specific variation and linguistic naturalness, a pattern that has also been observed in recent shared-task evaluations~\citep{atwany-etal-2024-osact,robinson-etal-2023-chatgpt}.

Motivated by these findings, we evaluate the impact of machine translation training on diglossia and dialect fidelity, following \textbf{Q1}.
\paragraph{Dialectal Text Generation with LLMs}
%
%
Instruction fine-tuning has demonstrated strong performance in controllable text generation~\citep{liang-etal-2024-controllable}. In dialectal settings, explicit instruction conditioning and the use of dialect tokens have been shown to improve alignment between generated outputs and target dialectal varieties~\citep{barmandah-2025-saudi}. However, prior work indicates that such approaches may also introduce generation artifacts when control constraints are emphasized, with potential negative effects on semantic fidelity~\citep{zhang-etal-2023-survey}.
We therefore evaluate the impact of instruction fine-tuning using next-word prediction on diglossia and dialect fidelity in the context of Arabic dialects (\textbf{Q2}).

%
\paragraph{Combining the Two Tasks}
Despite advances in both tasks, combining translation-based and generation-based techniques for DA remains underexplored. Prior research on Arabic NLP has begun to explore multi-task learning paradigms, for example, joint modeling of dialect identification and translation to improve MT quality~\citep{khered-etal-2025-multi}. There have also been efforts in other settings (e.g. unsupervised MT) to couple translation objectives with language modeling to better preserve fluency~\citep{artetxe-etal-2018-unsupervised}. Yet, to our knowledge, no prior work has explicitly jointly optimized a large language model for translation and dialect generation in the Arabic diglossia context. This gap motivates our approach to combine MT and next-word completion (\textbf{Q3}).
\section{Methodology}
\label{sec:methodology}
In this section, we go through the evaluation protocol (Subsection~\ref{subsec:evaluation}) and the training objectives (Subsection~\ref{subsec:training_objectives}).
\subsection{Evaluation Protocol}
\label{subsec:evaluation}
We followed the evaluation framework proposed by \citet{robinson-etal-2025-al} to assess two complementary dimensions of DA generation: fidelity and diglossia.

\paragraph{Fidelity}
This dimension is evaluated using both monolingual and cross-lingual prompts, in which the model is instructed to generate text in a specific DA variety. In such settings, no single gold reference exists, as multiple valid outputs may correspond to the same prompt. Accordingly, fidelity is measured using the Macro ADI2 dialect fidelity score, which assesses whether the generated output is both dialectal and identifiable as the target variety.

\paragraph{Diglossia}
Diglossia evaluates the model’s ability to translate between MSA and DA, reflecting its capacity to differentiate between dialects and MSA. This dimension is assessed through bidirectional translation tasks (MSA$\rightarrow$DA and DA$\rightarrow$MSA). For these tasks, reference translations are available and performance is measured using ChrF++~\citep{popovic-2015-chrf}.

Together, these two dimensions provide a complementary evaluation of dialectal Arabic generation: fidelity focuses on adherence to the target dialect in open-ended settings, while diglossia assesses controlled meaning under reference-based translations. 

For each evaluation dimension, we compute the mean score over the different datasets. Specifically, machine translation (corpus-level) and fidelity (sentence-level) are each evaluated on their own dedicated datasets.
%
\subsection{Training Objectives}
\label{subsec:training_objectives}
%

The two training objectives previously mentioned in section~\ref{sec:introduction}  differ not in their optimization procedure but in the constraints imposed by the task formulation. Translation provides a reference to enforce meaning preservation with respect to a source sentence, resulting in a constrained output space. In contrast, dialectal generation is open-ended: for a given instruction and prefix, multiple continuations may be equally valid as long as they are the target dialect (or have linguistic similarities). 

Formally, let $\mathcal{D}_{\text{MT}}$ and $\mathcal{D}_{\text{GEN}}$ denote the instruction-formatted datasets used for translation and dialectal generation, respectively. All training examples are represented using an instruction-based format and optimized with a causal language model objective, where only assistant tokens contribute to the loss. The final training objective minimizes a weighted combination of losses over the two datasets:
\begin{equation}
\mathcal{L}_{\text{joint}}(\theta)
=
\lambda \, \mathrm{E}_{\mathbf{z}\sim\mathcal{D}_{\text{MT}}}\!\left[\mathcal{L}(\mathbf{z})\right]
+
(1-\lambda) \, \mathrm{E}_{\mathbf{z}\sim\mathcal{D}_{\text{GEN}}}\!\left[\mathcal{L}(\mathbf{z})\right]
\end{equation}
where $\lambda \in [0,1]$ controls the relative contribution of translation and generation supervision.

We evaluate the model with $\lambda \in \{0, 0.5, 1\}$. When $\lambda = 0$, the task corresponds to pure generation, whereas when $\lambda = 1$, it corresponds to a MT task.


\begin{table*}[!ht]
\centering
\small
\begin{tabularx}{\textwidth}{p{3.8cm} X p{5.6cm}}
\toprule
\textbf{Instruction type} &
\textbf{Template} &
\textbf{Completion} \\
\midrule

English instruction &
\textit{Complete the sentence starting with these 3 words in <TARGET DIALECT>: <PREFIX>}
&
\textit{This is the full sentence in <TARGET DIALECT>: <TARGET TEXT IN DIALECT>} \\

\midrule
Dialectal instruction &
\parbox[t]{\linewidth}{
\begin{RLtext}
كمّل الجملة وابدأ بالتلات كلمات دول باللهجة
\LR{\texttt{<PREFIX>}}:
\LR{\texttt{<TARGET DIALECT>}}
\end{RLtext}
}

&
\parbox[t]{\linewidth}{
\begin{RLtext}
دي الجملة كاملة باللهجة
\LR{\texttt{<TARGET DIALECT>}}:
\end{RLtext}
} \\

\bottomrule
\end{tabularx}

\caption{Templates used for dialectal sentence completion in MADAR training data. Both templates require the model to generate a complete sentence in the target dialect starting from a fixed prefix; only the language of the instruction differs.}
\label{tab:generation_templates}
\end{table*}

\subsubsection{Machine Translation}
\label{subsec:machine_translation}
For the translation objective, each training example consists of an instruction in English specifying the translation direction (e.g.\ MSA$\rightarrow$DA, DA$\rightarrow$MSA, or DA$\leftrightarrow$English), followed by an assistant response containing the target sentence. After applying the chat template, the model is trained to maximize the conditional likelihood of the assistant tokens given the full preceding context. The template for this task is shown in Table~\ref{tab:mt_template}.


\begin{table}[t]
\centering
\small
\adjustbox{width=\columnwidth}{
\begin{tabular}{p{5.2cm} p{4.0cm} p{3.6cm}}
\toprule
\textbf{Instruction + Source} &
\textbf{Target (Reference or Output)} &
\textbf{Language Pair} \\
\midrule

\textit{Translate from <SRC> into <TGT>.}\\
<SOURCE SENTENCE> & <TARGET SENTENCE> & <SRC> $\rightarrow$ <TGT> \\ Translation: \\

\bottomrule
\end{tabular}
}
\caption{Template for instructing the translation across Arabic varieties and English.}
\label{tab:mt_template}
\end{table}

Prompt examples for translation are illustrated in Table~\ref{app:tab:mt_examples}.
\subsubsection{Instruction Next-Token Generation}
\label{subsec:instruct_next_token_generation}
For dialectal generation, training examples are formulated as instruction-conditioned sentence completion tasks. Each example provides an explicit instruction specifying the target dialect, followed by the first three words of a sentence, which serve as a fixed prefix. The model is trained to generate a complete sentence in the target dialect, starting from this prefix.

Two instruction variants are used. In the first variant, the instruction is formulated in English and specifies both the completion task and the target dialect. In the second variant, the instruction is formulated directly in the target dialect. In both cases, the assistant response contains a full sentence that repeats the provided prefix and continues it in a linguistically coherent manner. Although reference continuations are provided during training, they are not assumed to be unique, as the task may admit multiple valid dialectal realizations for the same prefix. Thus, the training signal encourages the model to learn distributional properties of the target dialect rather than to reproduce a single fixed continuation. The generation templates are provided in Table~\ref{tab:generation_templates} and examples are shown in Table~\ref{app:tab:generation_examples}.
\section{Experimental Set-Up}
\label{sec:experimental_set-up}
\subsection{Models}
\label{sec:models}
After a hyper-parameter search, 
we selected SmolLM3-3B to carry out our experiments \citep{bakouch2025smollm3}\footnote{\url{https://huggingface.co/HuggingFaceTB/SmolLM3-3B}} as it offers a good balance between model size, performance across tasks, and has been trained with Arabic data.

We instruction fine-tuned SmolLM3-3B to support multiple training objectives and evaluation regimes, selecting the best model according to both MT and next-token generation performance. Periodic evaluation was performed every 1,000 steps using the validation loss and a character-based metric (ChrF++). This design enables a direct comparison of models trained on (a) MT only, (b) MT combined with next-token generation, and (c) next-token generation only, while keeping the evaluation procedure consistent across experimental conditions. We let the reader refer to the model training in Appendix~\ref{subsec:instruct-fine-tuning}.

To address the research questions, we compare these training configurations against a baseline. The baseline corresponds to SmolLM3-3B with a hyperparameter search over decoding settings ($top$-$p \in \{0.1, 0.3, 0.6, 0.9, 1\}$ and $temperature \in \{0.1, 0.3, 0.6, 0.9, 1\}$). We shorten the original template by one third to train our instruction models. We found that both templates provided similar results in preliminary evaluations.

To assess the effect of scaling to a larger model, we replicate the same experimental setting using Llama-3.1-8B-Instruct\footnote{\url{https://huggingface.co/meta-llama/Llama-3.1-8B-Instruct}} as the base model, fine-tuned with LoRA (see Appendix~\ref{subsec:instruct-fine-tuning}). SmolLM-3B builds upon the Llama architecture, with modifications optimized for efficiency and long-context performance, which makes it a suitable point of comparison with Llama-3.1-8B-Instruct. We additionally compare our approach with a larger model to assess its effectiveness relative to models of different sizes, using the best configuration identified during hyperparameter search (refer to Appendix~\ref{app:hyperparameter_search}).
\subsection{Evaluation Data}
We adopted the same evaluation data and protocol described in Al Qasida  ~\cite{robinson-etal-2025-al}. Our evaluation set comprises both monolingual and crosslingual prompts across multiple DA varieties, designed to support text generation and MT tasks.
The crosslingual prompts were drawn from three different collections of LLMs user inputs: (i) a subset of Okapi prompts ~\cite{lai-etal-2023-okapi} used with the Alpaca LLM ~\cite{taori2023alpaca}, (ii) a collection of ChatGPT prompts obtained via the ShareGPT API, and (iii) a set of human-curated prompts by ~\citet{marchisio-etal-2024-understanding}.
In addition, our evaluation corpus incorporates both monolingual and bitext sentences from the corpus-6-test-corpus-26-dev split of the MADAR26 ~\cite{bouamor-etal-2018-madar}, a large-scale parallel resource covering dialects from seven Arab countries and consisting of BTEC-style everyday utterances originally introduced by~\citet{takezawa-etal-2007-multilingual}. We further included data from the FLORES200 dev, a multilingual benchmark based on manually translated Wikipedia text, selecting dialectal Arabic subsets representing five major regional varieties  ~\cite{nllb2022flores}. Finally, we integrated dialectal Arabic song lyrics from the HABIBI corpus, which spans eight Arab country dialects ~\cite{elhaj2020habibi}.
\subsection{Training Data}
\label{subsec:training_data}
Following the closed-data track,\footnote{\url{https://sites.google.com/view/vardial-2026/shared-tasks}} we used two sources of training data: bilingual data for task (i and ii, refer to Section~\ref{subsec:machine_translation} and~\ref{subsec:instruct_next_token_generation} respectively) and monolingual data for task (ii). 
\paragraph{Bilingual Data} 
Our bilingual training data were used to support translation tasks between English, MSA, and DA. For Saudi Arabic, the SauDial corpus ~\cite{ALANAZI2025111906} provided parallel data for EN$\leftrightarrow$DA and DA$\leftrightarrow$MSA translation. Palestinian Arabic–English parallel data were sourced from the Casablanca corpus  ~\cite{talafha2024casablanca}. For Jordanian Arabic, the JODA corpus  ~\cite{abandah2025joda} was used, offering parallel data between dialectal text and its MSA-corrected version. Syrian Arabic bilingual resources included the UFAL parallel corpus~\cite{krubinski-etal-2023-multi}, covering MSA$\leftrightarrow$DA and DA$\leftrightarrow$EN translation directions. Moroccan Arabic bilingual data combined several sources: the DODA corpus ~\cite{outchakoucht2024evolution} for EN$\leftrightarrow$DA translation, and the Atlas training sets ~\cite{atlasia2025atlasetblog}. These data were used to create the machine translation training data (see Section~\ref{subsec:machine_translation}).

\paragraph{Monolingual Data}
The monolingual training data were compiled from multiple resources covering a wide range of DA varieties. Saudi Arabic monolingual data were obtained from the SDC corpus~\cite{tarmom_codeswitching_arabic} as well as from the Saudi Tweets Corpus ~\cite{Alruily2018}. The Shami corpus~\cite{abu-kwaik-etal-2018-shami} was used to provide monolingual data for Palestinian, Syrian, and Jordanian varieties, while the MASC corpus~\cite{e1qb-jv46-21} contributed data for Egyptian and Jordanian Arabic. Moroccan Arabic monolingual data were sourced from the Goud training set~\cite{aftiss2025empirical}. Egyptian Arabic monolingual data were enriched using the EDGAD corpus~\cite{article,HUSSEIN2019109}, the EDC corpus~\cite{tarmom_codeswitching_arabic}, and the ASR-EgAr corpus~\cite{asr_egarbcsc}.

We also incorporated multidialectal monolingual data from the MADAR training set, which includes additional Arabic dialects. These data were used to create the instruction next-token generation data (see Section~\ref{subsec:instruct_next_token_generation}).

\section{Results}
\label{sec:results}

\begin{figure}
    \centering
    \includegraphics[width=1\linewidth]{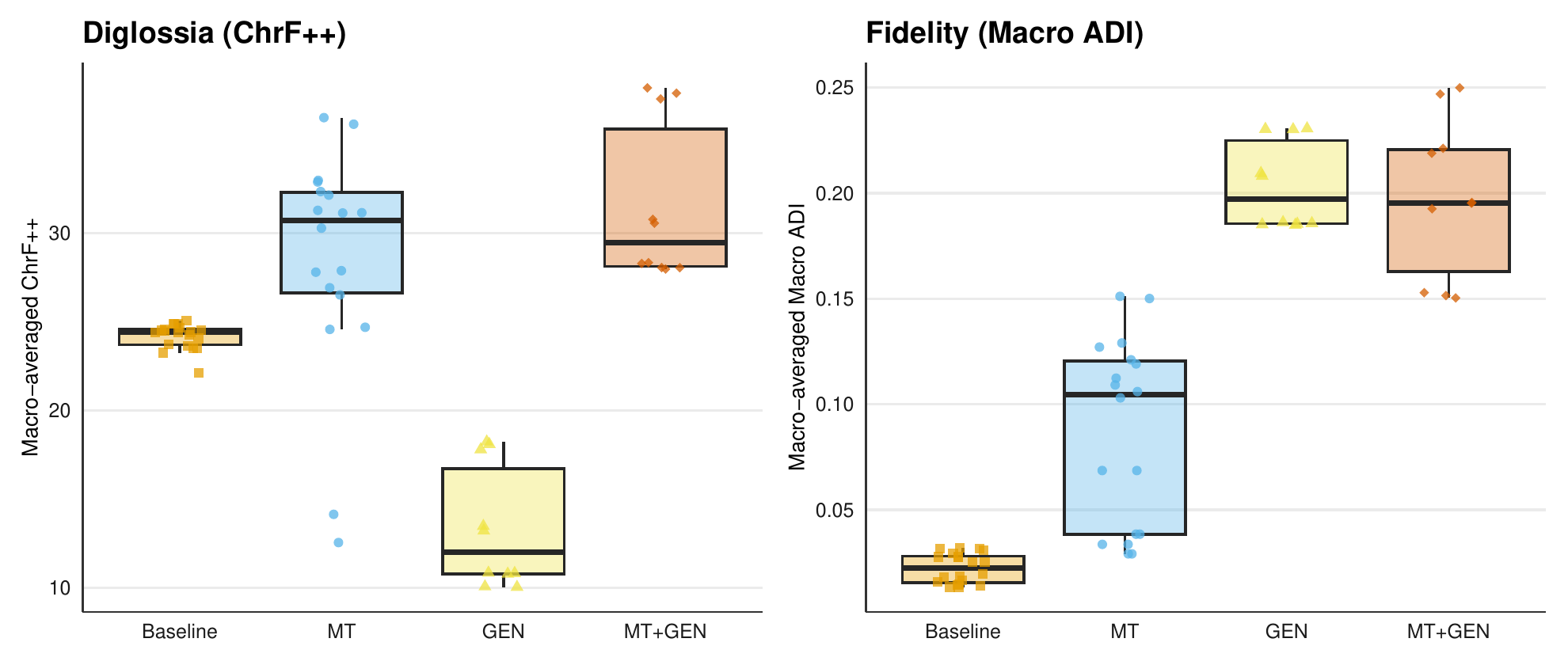}
    \caption{Performance for diglossia (ChrF++) and fidelity (Macro ADI2) across training paradigms. Each boxplot corresponds to a training paradigm (Baseline, MT, GEN, MT+GEN) using SmolLM3-3B, and each point represents a distinct decoding configuration (top\_p, temperature, learning rate), with scores macro-averaged over language varieties and test sets.}
    \label{fig:tasks_vs_test_set}
\end{figure}

%
\paragraph{Effect of the Task}
Figure~\ref{fig:tasks_vs_test_set} shows the scores with the different training objectives for the diglossia and fidelity tasks. Training with a translation objective improves diglossic scores, as reflected by higher ChrF++ across configurations (\textbf{Q1}). While MT also increases dialectal fidelity compared to the baseline, the gains remain limited and highly variable, suggesting that the LLM does not consistently generate the target dialect according to the Macro ADI2 score.

Instruction next-token generation improves dialectal fidelity, yielding the highest Macro ADI2 scores with low variance across configurations. However, this comes at the cost of diglossia, as generation-only models perform poorly on translation tasks, indicating a semantic drift (answering our \textbf{Q2}).

Jointly optimizing translation and instruction-based generation objectives yields a balance between diglossic scores and dialectal fidelity. Compared to MT training, the joint model substantially improves Macro ADI2 without degrading ChrF++ (avoiding the semantic shift observed in generation-only models). This indicates that translation and generation objectives provide complementary supervision signals for modeling Arabic dialects. 

\begin{table*}[!ht]
\centering
\small
\begin{tabular}{lcc}
\toprule
\textbf{Model} & \textbf{Diglossia~$\uparrow$} & \textbf{Fidelity~$\uparrow$} \\
\midrule
SmolLM3-3B 
  & 22.23 & 0.003 \\
\quad + Machine Translation + Instruction (\textbf{Constrastive})
  & 33.65 & 0.067 \\
\midrule
Llama-3.1-8B-Instruct 
  & 32.99 & 0.065 \\
\quad + Machine Translation + Instruction (\textbf{Primary})
  & 35.09 & 0.233 \\
\midrule
Command R Arabic 
  & 46.60 & 0.053 \\
GPT-OSS-120B 
  & 47.82 & 0.237 \\
\bottomrule
\end{tabular}
\caption{Comparison of baseline LLMs in terms of diglossia (ChrF++) and dialectal fidelity (Macro ADI2) across all language. For each model, scores correspond to the best decoding configuration selected across temperature and top-$p$ sampling settings.}
\label{tab:llm_comparison}
\end{table*}reprint

\paragraph{Trade-off Between the Tasks}
Figure~\ref{fig:trade-off_mt_fidelity} illustrates the trade-off between diglossia (measured by ChrF++) and dialectal fidelity, measured by Macro ADI2, across all decoding configurations. Faded points represent individual decoding configurations, highlighting the variability induced by learning rate and checkpoint selection (also decoding hyper-parameters for the baseline), while the highlighted points correspond to the best-performing configuration selected per model.

The baseline model clusters in the lower-left region of the plot, exhibiting both limited diglossia and weak dialectal fidelity. Models trained exclusively with a MT objective achieve higher ChrF++ scores, indicating stronger diglossia, but remain constrained in dialectal fidelity. In contrast, instruction-based next-token generation (GEN) prioritizes dialectal generation, yielding higher Macro ADI2 scores at the cost of reduced diglossia scores.

The combined MT+GEN model is in the upper-right region of the plot, demonstrating that jointly optimizing translation and instruction-based generation objectives leads to a more favorable balance between semantic adequacy and dialectal expressiveness (\textbf{Q3}). %

\paragraph{Comparison with Other LLMs}
To better understand the impact of joint training, we also carried out the same experiments using Llama-3.1-8B-Instruct\footnote{\url{https://huggingface.co/meta-llama/Llama-3.1-8B-Instruct}} as the base model, fine-tuned with LoRA.

In addition, we included a 120B-parameter model\footnote{\url{https://huggingface.co/openai/gpt-oss-120b}} as a reference point to assess how performance scales with substantially larger model parameters (without any optimization method). We also considered Command R Arabic, a model specifically optimized for translation across multiple Arabic varieties\footnote{\url{https://huggingface.co/CohereLabs/c4ai-command-r7b-arabic-02-2025}}, in order to compare our approach against a system designed for Arabic multilingual translation. Taken together, these baselines provide an approximate upper bound on the performance and help contextualize the results of smaller jointly trained models.

Table~\ref{tab:llm_comparison} reports the performance of jointly trained models in comparison with their base model. The automatic evaluation scores indicate that, joint training across tasks leads to consistent improvements in metrics reflecting both diglossia and fidelity scores even for larger model (Llama-3.1-8B-Instruct) with LoRA fine-tuning. These results provide further evidence in support of the benefits of jointly training the model with both task (\textbf{Q3}).

Compared to the other models, the much larger GPT-OSS-120B attains the highest scores overall. Despite being substantially smaller, Llama-3.1-8B-Instruct reaches the fidelity score (Macro ADI2) of 0.23, achieving performance comparable to the 120B model. This suggests that dialectal control can be enhanced through supervision strategies rather than model scaling alone.
\section{Conclusion}
\label{sec:conclusion}
This work presented the submission of team \textbf{\mbox{Aladdin-FTI ~\includegraphics[height=1em]{man-genie.pdf}}} to the AMIYA shared task, aiming to model Arabic dialects through a unified framework that combines translation and instruction-based generation. 


Our results show that these objectives provide complementary supervision: translation improves diglossic awareness and semantic adequacy (\textbf{Q1}), while instruction-conditioned generation enhances dialectal fidelity (\textbf{Q2}). By combining both objectives, we obtain a more balanced model that consistently outperforms single-objective approaches across evaluation dimensions (\textbf{Q3}).

Notably, this balance is achieved with a smaller model that competes with larger systems, underscoring the importance of training objective design in dialectal Arabic modeling. These findings support treating Arabic as a pluricentric language.

Future work will focus on refining the balance between objectives, expanding coverage to additional varieties, and conducting human and linguistic evaluations to better assess dialectal naturalness.
\section*{Limitations}
\label{sec:limitations}
This study is limited to a single model architecture. While the results are encouraging, further experiments on different model families and scales are needed to assess the generality of the proposed approach. While few-shot and in-context learning approaches may be effective (see e.g. \citet{gao-etal-2021-making}; or \citet{mutal-etal-2025-factors} for use in low-resource settings), they were not considered in this work, as our objective was to keep input prompts compact and limit the number of tokens provided to the model.
\section*{Acknowledgments}
\label{sec:acknowledgments}
The computations were performed at the University of Geneva using the Baobab HPC service.

\bibliography{latex/custom, latex/arxiv_refs}

\appendix

\section{LLM Settings and Results}
\label{sec:appendix}

\subsection{Instruct Fine-Tuning}
\label{subsec:instruct-fine-tuning}
We fine-tuned two instruction-tuned models: SmolLM3-3B and Llama-3.1-8B-Instruct. Both models shared the same training configuration. Evaluation and checkpointing were performed every $1{,}000$ steps, and the best-performing model was retained according to ChrF++ and perplexity. ChrF++ evaluation relied on deterministic text generation (temperature 0.0, top-$p$ 1.0) with up to 512 generated tokens, and translation quality was assessed using ChrF++ with a character n-gram size of 64. The best checkpoint was selected by maximizing the macro-averaged ChrF++ score over the full development set.

All models were trained for four epochs with a per-device batch size of 16, gradient accumulation over eight steps (effective batch size 128), and a per-device evaluation batch size of eight. Optimization relied on AdamW with a cosine learning-rate scheduler, a warm-up ratio of 3\%, weight decay of 0.01, and gradient clipping with a maximum norm of 1.0. Learning rate values were swept over $\{2\times10^{-5}, 3\times10^{-5}, 5\times10^{-5}, 6\times10^{-5}\}$. For reproducibility, we fixed the random seed to 42.

Due to GPU resource constraints, we adopted two different instruct fine-tuning strategies:

\paragraph{SmolLM3-3B}
We fine-tuned the HuggingFaceTB/SmolLM3-3B model using a custom template to ensure alignment with the training data. Training was conducted in bfloat16 precision with TF32 enabled and gradient checkpointing. No parameter-efficient fine-tuning or quantization was applied for this model.

\paragraph{Llama-8B-Instruct}
We fine-tuned meta-llama/Meta-Llama-3.1-8B-Instruct using parameter-efficient adaptation with LoRA~\citep{hu-etal-2021-lora}. LoRA was applied with rank $r=16$, scaling factor $\alpha=32$, and dropout $0.05$, targeting the attention projection layers ($q$, $k$, $v$, $o$) and the feed-forward layers. No quantization was applied.
%

%
%
%
%
%
\clearpage
\onecolumn
\subsection{Machine Translation Template}
\label{app:subsection:machine_translation_template}

\begin{table*}[th!]
\centering
\small
\begin{tabularx}{\textwidth}{L R}
\toprule
\textbf{Instruction + Source (English)} &
\textbf{Reference Translation (Egyptian Arabic)} \\
\midrule

\textit{Translate from English into Egyptian Arabic. Output only the translation.}\\
I wonder if a table is available near the window for seven tonight.
&
\parbox[t]{\linewidth}{%
\begin{RLtext}
أنا بسأل لو كان فيه ترابيزة جنب الشباك تكون فاضية على الساعة سبعة بليل.
\end{RLtext}%
} \\

\midrule
\textit{Translate from English into Egyptian Arabic. Output only the translation.}\\
I feel chilly and my stomach hurts badly.
&
\parbox[t]{\linewidth}{%
\begin{RLtext}
حاسس إني بردان ومعدتي واجعاني جامد.
\end{RLtext}%
} \\

\midrule
\textit{Translate from English into Egyptian Arabic. Output only the translation.}\\
Can I invite you out for dinner some time?
&
\parbox[t]{\linewidth}{%
\begin{RLtext}
ممكن أعزمك على العشا في وقت ما؟
\end{RLtext}%
} \\

\midrule
\textit{Translate from English into Egyptian Arabic. Output only the translation.}\\
The very thing I was looking for.
&
\parbox[t]{\linewidth}{%
\begin{RLtext}
الحاجة اللي كنت بدوّر عليها.
\end{RLtext}%
} \\

\midrule
\textit{Translate from English into Egyptian Arabic. Output only the translation.}\\
Flight number two zero eight, to Tokyo.
&
\parbox[t]{\linewidth}{%
\begin{RLtext}
رحلة رقم ميتين وتمانية، لطوكيو.
\end{RLtext}%
} \\

\midrule
\textit{Translate from English into Egyptian Arabic. Output only the translation.}\\
I'd like a typewriter ribbon.
&
\parbox[t]{\linewidth}{%
\begin{RLtext}
عايز شريط آلة كاتبة.
\end{RLtext}%
} \\

\midrule
\textit{Translate from English into Egyptian Arabic. Output only the translation.}\\
What's the minimum age to do this?
&
\parbox[t]{\linewidth}{%
\begin{RLtext}
إيه أقل سن ممكن يعمل ده؟
\end{RLtext}%
} \\

\midrule
\textit{Translate from English into Egyptian Arabic. Output only the translation.}\\
Could you call me if you find my bag?
&
\parbox[t]{\linewidth}{%
\begin{RLtext}
ممكن تكلّمني لو لقيت شنطتي؟
\end{RLtext}%
} \\

\bottomrule
\end{tabularx}
\caption{Examples of instruction-conditioned machine translation from English to Egyptian Arabic (MADAR-26).}
\label{app:tab:mt_examples}
\end{table*}

\clearpage
\onecolumn
\subsection{Instruction Next-Token Generation}
\label{app:subsec:instruction_next-token_generation}
\begin{table*}[h]
\centering
\small
\begin{tabularx}{\textwidth}{l X X} 
\toprule
\textbf{Instruction language} &
\textbf{Instruction + Prefix} &
\textbf{Reference continuation (assistant)} \\
\midrule

English &
\textit{Complete the sentence starting with these 3 words in Egyptian Arabic:} \parbox[t]{\linewidth}{\begin{RLtext} ده قدامك هناك، \end{RLtext}}
& 
\parbox[t]{\linewidth}{\begin{RLtext} دي الجملة كاملة باللهجة المصرية: ده قدامك هناك، يادوبك قدام مكتب استعلامات السياحة. \end{RLtext}} \\

\midrule
Dialectal &
\parbox[t]{\linewidth}{\begin{RLtext} كمّل الجملة وابدأ بالتلات كلمات دول باللهجة المصرية: ده قدامك هناك، \end{RLtext}}
& 
\parbox[t]{\linewidth}{\begin{RLtext} دي الجملة كاملة باللهجة المصرية: ده قدامك هناك، يادوبك قدام مكتب استعلامات السياحة. \end{RLtext}} \\

\midrule
English &
\textit{Complete the sentence starting with these 3 words in Egyptian Arabic:} \parbox[t]{\linewidth}{\begin{RLtext} عمري ما سمعت \end{RLtext}}
& 
\parbox[t]{\linewidth}{\begin{RLtext} دي الجملة كاملة باللهجة المصرية: عمري ما سمعت عن العنوان ده هنا. \end{RLtext}} \\

\midrule
Dialectal &
\parbox[t]{\linewidth}{\begin{RLtext} كمّل الجملة وابدأ بالتلات كلمات دول باللهجة المصرية: عمري ما سمعت \end{RLtext}}
& 
\parbox[t]{\linewidth}{\begin{RLtext} دي الجملة كاملة باللهجة المصرية: عمري ما سمعت عن العنوان ده هنا. \end{RLtext}} \\

\bottomrule
\end{tabularx}
\caption{Examples of instruction-conditioned dialectal generation from MADAR training data. The task consists of completing a sentence in a target dialect from a fixed three word prefix. Instructions are provided either in English or directly in the target dialect, while the generation objective remains identical.}
\label{app:tab:generation_examples}
\end{table*}
%
%

\clearpage
\onecolumn
\subsection{Hyperparameter Search Results}
\label{app:hyperparameter_search}

In this section, we show the performance of each model under different decoding hyperparameter settings.

\begin{table}[!ht]
\centering
\small
\begin{tabular}{@{}S[table-format=1.1]S[table-format=1.1]|S[table-format=2.3]S[table-format=1.3]|S[table-format=2.3]S[table-format=1.3]|S[table-format=2.3]S[table-format=1.3]|S[table-format=2.3]S[table-format=1.3]@{}}
\toprule
\multicolumn{2}{@{}l}{\textbf{Hyperparameters}} & \multicolumn{2}{@{}l}{\textbf{SmolLM3-3B}} & \multicolumn{2}{@{}l}{\textbf{Llama-3.1-8B-Instruct}} & \multicolumn{2}{@{}l}{\textbf{Command R Arabic}} & \multicolumn{2}{@{}l}{\textbf{GPT-OSS-120B}}\\
\textbf{top-$p$} & \textbf{T} & \textbf{Diglossia} & \textbf{Fidelity} & \textbf{Diglossia} & \textbf{Fidelity} & \textbf{Diglossia} & \textbf{Fidelity} & \textbf{Diglossia} & \textbf{Fidelity} \\
\midrule
0.1 & 0.1 & 21.96 & 0.003 & 32.96 & 0.044 & 45.34 & 0.018 & 47.14 & 0.200 \\
0.1 & 0.3 & 22.00 & 0.003 & 32.96 & 0.044 & 45.35 & 0.019 & 47.39 & 0.206 \\
0.1 & 0.6 & 22.23 & 0.003 & 32.97 & 0.045 & 45.97 & 0.019 & 47.30 & 0.205 \\
0.1 & 0.9 & 22.00 & 0.003 & 32.95 & 0.043 & 46.42 & 0.020 & 47.17 & 0.227 \\
0.1 & 1   & 21.85 & 0.003 & 32.83 & 0.044 & 46.19 & 0.021 & 47.60 & 0.214 \\
0.3 & 0.1 & 21.96 & 0.003 & 32.97 & 0.044 & 46.42 & 0.020 & 47.50 & 0.207 \\
0.3 & 0.3 & 21.93 & 0.003 & 32.97 & 0.044 & 46.46 & 0.020 & 47.55 & 0.204 \\
0.3 & 0.6 & 21.83 & 0.003 & 32.86 & 0.043 & 46.47 & 0.021 & 47.49 & 0.224 \\
0.3 & 0.9 & 21.40 & 0.003 & 32.49 & 0.044 & 46.52 & 0.020 & 47.37 & 0.216 \\
0.3 & 1   & 20.96 & 0.003 & 32.20 & 0.047 & 46.60 & 0.053 & 47.32 & 0.237 \\
0.6 & 0.1 & 21.81 & 0.003 & 32.97 & 0.043 & 46.23 & 0.023 & 47.48 & 0.188 \\
0.6 & 0.3 & 21.80 & 0.003 & 32.91 & 0.045 & 46.19 & 0.022 & 47.53 & 0.210 \\
0.6 & 0.6 & 21.18 & 0.003 & 32.31 & 0.045 & 46.18 & 0.022 & 47.72 & 0.216 \\
0.6 & 0.9 & 18.96 & 0.002 & 30.82 & 0.046 & 46.12 & 0.023 & 47.13 & 0.208 \\
0.6 & 1   & 16.86 & 0.002 & 29.69 & 0.043 & 46.05 & 0.021 & 46.78 & 0.203 \\
0.9 & 0.1 & 21.85 & 0.002 & 32.97 & 0.045 & 45.19 & 0.021 & 47.45 & 0.209 \\
0.9 & 0.3 & 21.68 & 0.002 & 32.67 & 0.044 & 45.08 & 0.020 & 47.48 & 0.223 \\
0.9 & 0.6 & 18.91 & 0.003 & 30.81 & 0.049 & 45.13 & 0.021 & 47.39 & 0.215 \\
0.9 & 0.9 & 13.76 & 0.002 & 27.34 & 0.055 & 45.23 & 0.021 & 47.19 & 0.197 \\
0.9 & 1   & 11.40 & 0.002 & 25.21 & 0.058 & 45.33 & 0.020 & 46.83 & 0.211 \\
1   & 0.1 & 21.78 & 0.003 & 32.99 & 0.045 & 44.83 & 0.020 & 47.50 & 0.217 \\
1   & 0.3 & 21.27 & 0.002 & 32.54 & 0.045 & 44.75 & 0.020 & 47.82 & 0.213 \\
1   & 0.6 & 17.44 & 0.002 & 30.06 & 0.047 & 44.88 & 0.021 & 47.31 & 0.202 \\
1   & 0.9 & 12.09 & 0.002 & 25.70 & 0.048 & 44.72 & 0.020 & 46.97 & 0.219 \\
1   & 1   &  9.78 & 0.002 & 23.66 & 0.065 & 44.78 & 0.020 & 45.68 & 0.218 \\
\bottomrule
\end{tabular}
\caption{Decoding performance for different top-$p$ and temperature (T) settings, evaluated with Diglossia (ChrF++) and Fidelity (Macro ADI2).}
\label{app:tab:decoding_performance}
\end{table}

The table~\ref{app:tab:decoding_performance} reports the impact of varying decoding hyperparameters $p$ and temperature) on both diglossia (ChrF++) and dialectal fidelity (Macro ADI2) across several models. Overall, performance remains relatively stable for moderate decoding settings, where diglossia scores vary only slightly within each model. However, increasing temperature leads to more diverse but less controlled generation, which often results in degraded diglossia scores, particularly for smaller models such as SmolLM3-3B, whose ChrF++ drops sharply at higher temperature values ($T=1, top-p=1, diglossia=9.78$ diglossia score). OpenGPT-OSS-120B (the largest model) remain more robust, maintaining high diglossia and fidelity scores across most configurations.
\end{document}